\documentclass[10pt,twocolumn,letterpaper]{article}
\usepackage[T1]{fontenc}
\usepackage{times} 
\usepackage{cvpr}              
\usepackage{float}
\usepackage{algorithm}
\usepackage{algorithmic}
\usepackage{caption}
\usepackage{dsfont}
\usepackage{tabularx}
\usepackage{diagbox}
\usepackage[accsupp]{axessibility}
\usepackage{multirow}

\usepackage{makecell}
\usepackage{bbding} 
\usepackage[dvipsnames]{xcolor}

\definecolor{cvprblue}{rgb}{0.21,0.49,0.74}
\usepackage[pagebackref,breaklinks,colorlinks,citecolor=cvprblue]{hyperref}

\title{Positive2Negative: Breaking the Information-Lossy Barrier \\in Self-Supervised Single Image Denoising}

\def\authorBlock{
	Tong Li$^{1} $ \qquad Lizhi Wang$^{2,3}$\thanks{Corresponding Author: Lizhi Wang (wanglizhi@bnu.edu.cn)} \qquad Zhiyuan Xu$^{1}$ \qquad Lin Zhu$^{1}$ \qquad Wanxuan Lu$^{4}$ \qquad Hua Huang$^{2,3}$ \\
	$^{1}$ School of Computer Science and Technology, Beijing Institute of Technology \\
	$^{2}$ School of Artificial Intelligence, Beijing Normal University \\
	$^{3}$ Engineering Research Center of Intelligent Technology and Educational Application, Ministry of Education \\
	$^{4}$Aerospace Information Research Institute, Chinese Academy of Sciences\\}
	
\begin{document}
\author{\authorBlock}
\maketitle

\begin{abstract}
Image denoising enhances image quality, serving as a foundational technique across various computational photography applications. 
The obstacle to clean image acquisition in real scenarios necessitates the development of self-supervised image denoising methods only depending on noisy images, especially a single noisy image. 
Existing self-supervised image denoising paradigms (Noise2Noise and Noise2Void) rely heavily on information-lossy operations, such as downsampling and masking, culminating in low-quality denoising performance.
In this paper, we propose a novel self-supervised single image denoising paradigm, Positive2Negative, to break the information-lossy barrier.
Our paradigm involves two key steps: Renoised Data Construction (RDC) and Denoised Consistency Supervision (DCS). RDC renoises the predicted denoised image by the predicted noise to construct multiple noisy images, preserving all the information of the original image. DCS ensures consistency across the multiple denoised images, supervising the network to learn robust denoising. 
Our Positive2Negative paradigm achieves state-of-the-art performance in self-supervised single image denoising with significant speed improvements. 
The code is released to the public at \url{https://github.com/Li-Tong-621/P2N}.
\end{abstract}


\section{Introduction}

Image denoising plays a vital role in image processing to enhance visual quality~\cite{hasinoff2016photography}, serving as a foundational technique across various computational photography applications~\cite{wang2020photography,fluorescence_survey}.
The widespread popularity of paired clean-noisy image datasets~\cite{SIDD,DND} has catalyzed advancements in supervised methods, yielding remarkable performance~\cite{Restormer,NAFNet}.
However, supervised methods fall short in dynamic scenes~\cite{N2N,N2V}, 
where long exposure results in misalignment and blur~\cite{N2N,ELD}, hindering the 
acquisition of clean images.
Therefore, advancing self-supervised image denoising methods using only noisy images~\cite{CARE,DeepCADRT}, especially a single noisy image~\cite{MASH,DMID,ZSN2N,ScoreDVI}, holds significant importance.

\begin{figure}
	\centering
	\includegraphics[width=\linewidth]{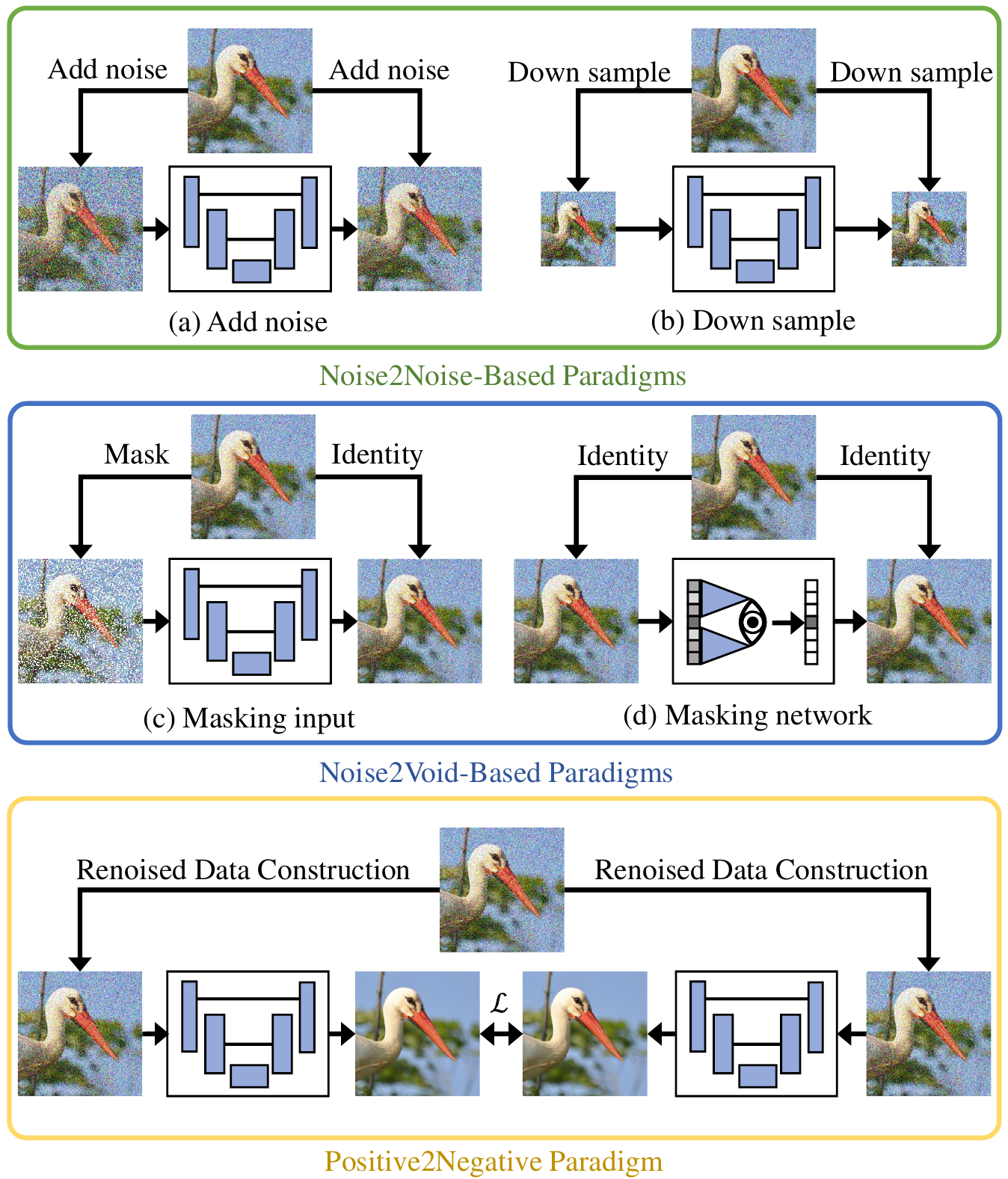} 
	\caption{\textbf{Information-lossy operations in current paradigms}. 
	The noise addition operation reduces the signal-to-noise ratio, the downsampling operation diminishes the sampling density, and the masking operation disregards crucial central pixels. 
	Each of these operations introduces an information-lossy barrier~\cite{ATBSN,Nyquist,Shannon}, culminating in imprecise denoising.
	}
	
	\label{fig:probelm}
	\vspace{-1em}
\end{figure}

The crux of self-supervised image denoising methods lies in the construction of the training data and the formulation of the supervision strategy.
Current methods are divided into Noise2Noise-based paradigms and Noise2Void-based paradigms, as illustrated in Figure~\ref{fig:probelm}. 
Noise2Noise-based paradigms add additional noise to the noisy image~\cite{R2R,Noisier2noise} or down sample the noisy image~\cite{Neighbor2neighbor,ZSN2N} to approximately construct two independently noisy observations as training data. 
These paradigms learn denoising by supervising one observation with the denoised result of other observation.
Noise2Void-based paradigms mask either the noisy image~\cite{Blind2Unblind,N2V} or the neural network~\cite{APBSN,LGBPN,ATBSN} to construct masked training data.
These paradigms learn denoising by supervising the denoised result of the masked central pixel with the original noisy pixel, 
where the prediction only relies on the surrounding pixels.
 
\begin{figure}[t]
	\centering
	
	\renewcommand{\arraystretch}{1} 
	\setlength{\tabcolsep}{1.5pt}
	\scalebox{0.97}{
		\small
		\begin{tabular}{c c c c}
			
			\includegraphics[width=.115\textwidth]{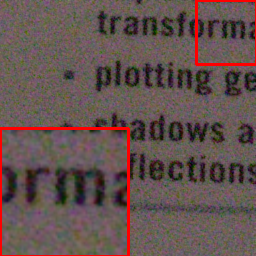} &   
			\includegraphics[width=.115\textwidth]{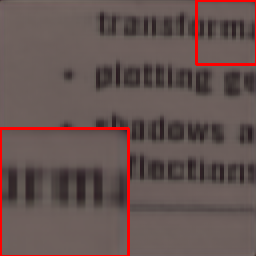} &
			\includegraphics[width=.115\textwidth]{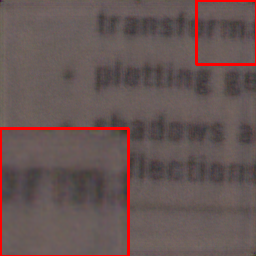} &
			\includegraphics[width=.115\textwidth]{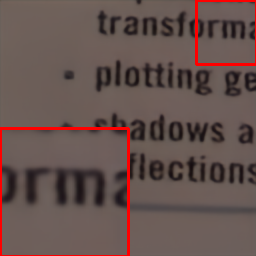}
			
			\\
			\includegraphics[width=.115\textwidth]{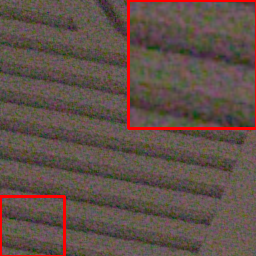} &   
			\includegraphics[width=.115\textwidth]{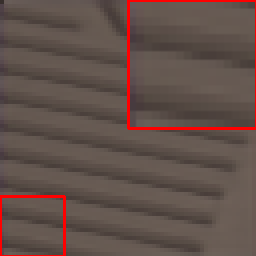} &
			\includegraphics[width=.115\textwidth]{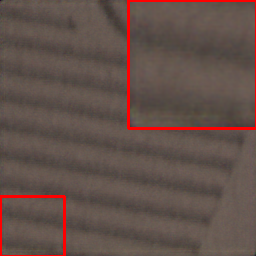} &
			\includegraphics[width=.115\textwidth]{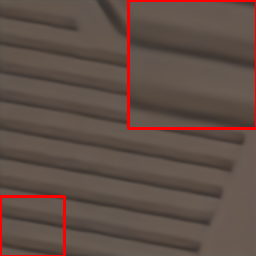}
			
			\\
			ZS-N2N~\cite{ZSN2N} &
			MASH~\cite{MASH} &
			ATBSN~\cite{ATBSN} &
			\textbf{Ours}
						
			\\			
			
		\end{tabular}
	}
	\caption{
		\textbf{The information-lossy barrier significantly compromises the image quality}, resulting in residual noise, aliasing effects or texture loss.
	}
	\label{fig:performance}
		\vspace{-1em}
\end{figure}

However, existing methods rely heavily on information-lossy operations, 
resulting in low-quality denoising performance~\cite{Blind2Unblind,SDAP,LGBPN,APBSN}. 
Specifically, the noise addition operation reduces the signal-to-noise ratio, the downsampling operation diminishes the sampling density and the masking operation disregards crucial central pixels. 
Each of these operations loses the information from the original noisy image~\cite{ATBSN,Nyquist,Shannon}, 
inevitably compromising the denoising ability~\cite{learnability,LGBPN}.
Consequently, denoised images always suffer from remaining noise, aliasing effects or texture loss~\cite{PPDP,PLMIC}, as shown in Figure~\ref{fig:performance}.
As these information-lossy operations constitute the indispensable foundation of the Noise2Noise-based and Noise2Void-based paradigms, 
it is imperative to establish a new paradigm that transcends the Noise2Noise-based and Noise2Void-based paradigms.

In this paper, we propose the concise Positive2Negative paradigm to break the information-lossy barrier in self-supervised single image denoising.
Our inspiration stems from the observation that noise distributions are approximately symmetrical and centered around zero, implying that the opposite noise also follows the distribution of the original noise. 
Leveraging this insight, we can construct multiple noisy images corresponding to the same clean image, 
which serves as the basis of Positive2Negative.
Concretely, Positive2Negative relies on two steps: Renoised Data Construction (RDC) and Denoised Consistency Supervision (DCS).

The basic idea of RDC is to construct training data.
Specifically, we design a zero-mean sampling strategy to generate multi-scale Positive noise and Negative noise based on the predicted noise, which are then added back to construct Positive noisy images and Negative noisy images. 
The constructed data comprehensively covers the information of the given single image; thus the Positive2Negative paradigm breaks the information-lossy barrier.

The basic idea of DCS is to formulate a supervision strategy. 
Specifically, given the constructed multi-scale noisy images, DCS supervises the network to generate consistent denoised outputs during the training process. 
Furthermore, we provide a theoretical analysis showing that the Positive2Negative paradigm can learn robust denoising, through supervising the multiple denoised images.

In summary, our contributions are as follows:

\begin{itemize}[leftmargin=0.8cm]

	\item We propose a new self-supervised single image denoising paradigm, Positive2Negative, to break the information-lossy barrier.

	\item We propose a data construction method, which constructs multi-scale similar noisy images for training.
	
	\item We propose a denoising supervision method, which is theoretically guaranteed to learn robust denoising.

	\item Positive2Negative achieves SOTA performance over self-supervised single image denoising.
\end{itemize}

\section{Related Work}
From the training data perspective, the self-supervised image denoising methods can be classified into noisy dataset based methods and single image based methods.


\subsection{Noisy Dataset based Methods}


{Noisy dataset} based methods represent the most active research direction in image denoising. These methods learn to denoise from a given noisy dataset. Generally, a larger volume of the noisy dataset improves the denoising performance. As a result, in scenarios where only a single noisy image is available, the denoising performance of whole noisy dataset-based methods decreases significantly.

Noise2Noise~\cite{N2N} learns denoising by mapping one noisy observation to another independent observation of the same scene. 
Subsequent studies focus on generating observations of the same scene from a single observation, typically achieved through downsampling~\cite{Neighbor2neighbor,ZSN2N} or by introducing additional noise~\cite{Noisier2noise,R2R,NAC,IDR,DCD}.
After that, another elegant method Noise2Void~\cite{N2V} is proposed. Noise2Void~\cite{N2V} assumes that the noise distribution is zero-mean, the noise is spatially independent, and the signal is spatially correlated.
Noise2Void~\cite{N2V} predicts the clean signal by leveraging surrounding noisy signals, with masking the central pixel. 
Various other works propose alternative masking schemes, either mask the input~\cite{Blind2Unblind,Self2self,MASH} or mask the network~\cite{Noise2self,Laine,Noise2Same,MMBSN,PN2V,SwinIA}.
Noise2Void struggles with real-world noise, which often exhibits spatial correlation.
Therefore, subsequent research endeavors to break the noise spatial correlation~\cite{PD,D-BSN,APBSN,LGBPN,ATBSN,li2023spatially}. 
There are also some methods amalgamate frameworks established by both Noise2Noise and Noise2Void ~\cite{SDAP,CBSD} or get rid of frameworks established by Noise2Noise and Noise2Void~\cite{FBI,CVF-SID,LUD-VAE,NDA,Noise2Score}, but these methods are usually less developed due to the complexity. 

{Noisy dataset} based methods have high requirements for the volume of training data and fail in scenarios where only a single noisy image is available~~\cite{MASH,ScoreDVI}.

\subsection{Single Image based Methods}
Single image based methods represent the most challenging situation for denoising and the least demanding requirements for data. These methods learn to denoise from a given single noisy image without any dependence on paired noisy-clean or noisy datasets. As a result, single image based methods are less developed due to the inherent difficulty, only gaining attention in the past two years.

DIP~\cite{DIP} is the earliest self-supervised image denoising method. DIP~\cite{DIP} learns to map random noise to a given noisy image and employs early stopping to prevent overfitting. 
Self2Self~\cite{Self2self} uses dropout to create two Bernoulli-sampled observations of a noisy image. 
R2R~\cite{R2R} constructs noisy images by adding new Gaussian noise and applying reciprocal coefficients.
ScoreDVI~\cite{ScoreDVI} performs real denoising using a pre-trained Gaussian denoiser and an estimated noise model.
ZS-N2N~\cite{ZSN2N} extend Noise2Noise~\cite{N2N} to a single image version by downsampling.
MASH~\cite{MASH} employs masking and local pixel shuffling to break the noise spatial correlation.
Recently, DMID stimulates the diffusion model for self-supervised single image denoising~\cite{DMID}.

In this paper, we propose Positive2Negative, a new self-supervised single image denoising paradigm, to break the information-lossy barrier.

\section{Positive2Negative}
\label{sec:method}
In this section, we first demonstrate the motivation of Positive2Negative in Section~\ref{sec:Motivation}. After that, we present the two critical steps of Positive2Negative: RDC and DCS, respectively in Section~\ref{sec:RDC} and Section~\ref{sec:DCS}. Finally, we provide an analysis and proof of Positive2Negative in Section~\ref{sec:Analysis}.

\begin{figure}
	\centering
	\vspace{-1em}
	\includegraphics[width=\linewidth]{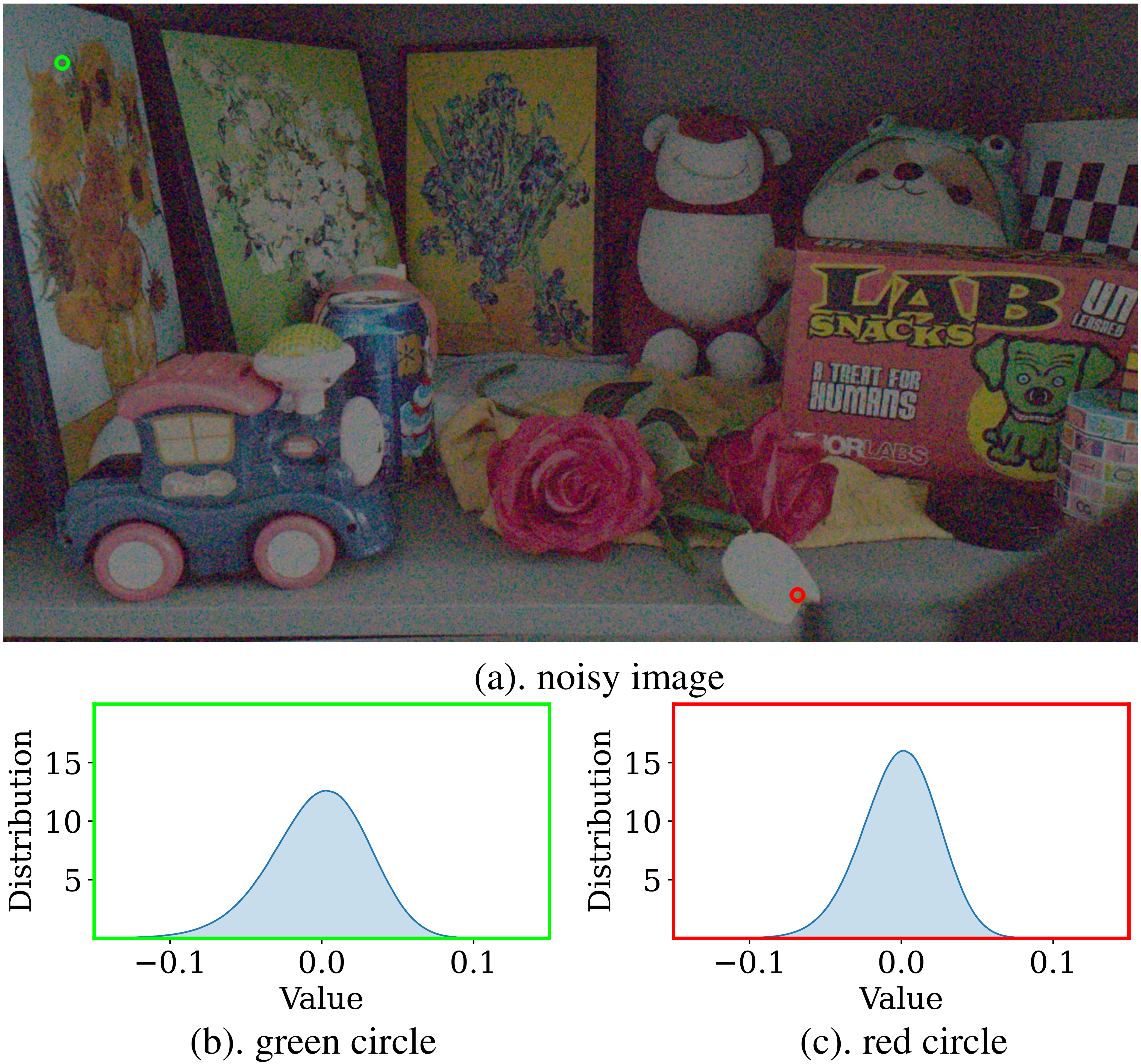} 
	\caption{\textbf{Noise distribution is zero-mean and approximately symmetrical}. 
(a) shows a noisy image. (b) and (c) show the noise distributions, which are calculated at the center pixels marked by the {\color{green}green circle} and the {\color{red}red circle} in the noisy image, respectively. 
It is evident that the noise distribution is zero-mean and approximately symmetrical.
	}
	\label{fig:noise_distribution}
\end{figure}

\begin{figure}[t]
	\centering
	
	\renewcommand{\arraystretch}{1} 
	\setlength{\tabcolsep}{1.5pt}
	\scalebox{0.97}{
		\small
		\begin{tabular}{c c c c}
			
			\includegraphics[width=.115\textwidth]{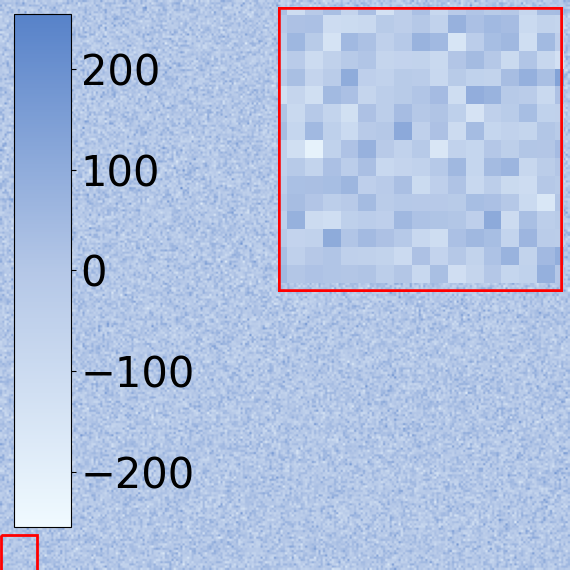} &   
			\includegraphics[width=.115\textwidth]{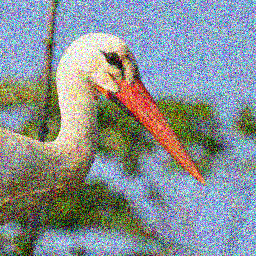} &
			\includegraphics[width=.115\textwidth]{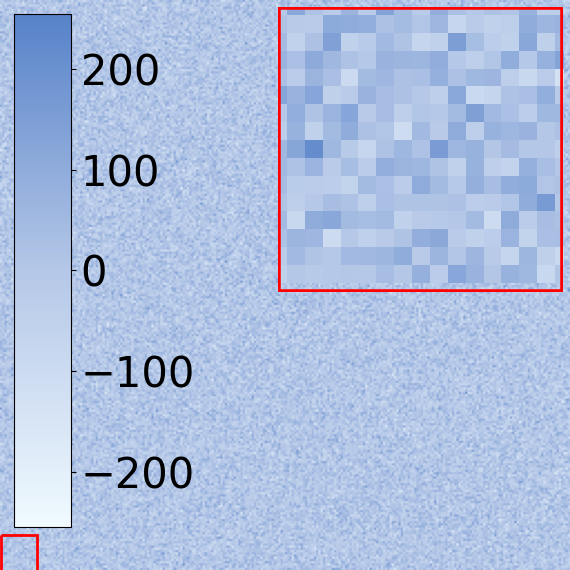} &
			\includegraphics[width=.115\textwidth]{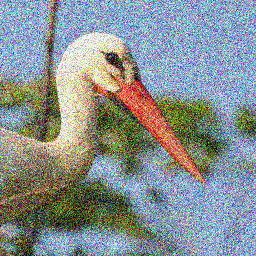}	
			\\
			Original noise &
			Original noisy &
			Opposite noise &
			Opposite noisy
			\\
			$n$ &
			$y = c + n$ &
			$n_n = -n$ &
			$y_n = c + (-n)$	
			\\
			
		\end{tabular}
	}
	\caption{
	The opposite noise $-n$ is similar to the original noise $n$.
	The opposite noise $-n$ approximately follows the original noise's distribution, which is zero-mean and approximately symmetrical.
	}
	\label{fig:intuitive}
		\vspace{-1em}
\end{figure}

\begin{figure*}
	\centering
	\includegraphics[width=\linewidth]{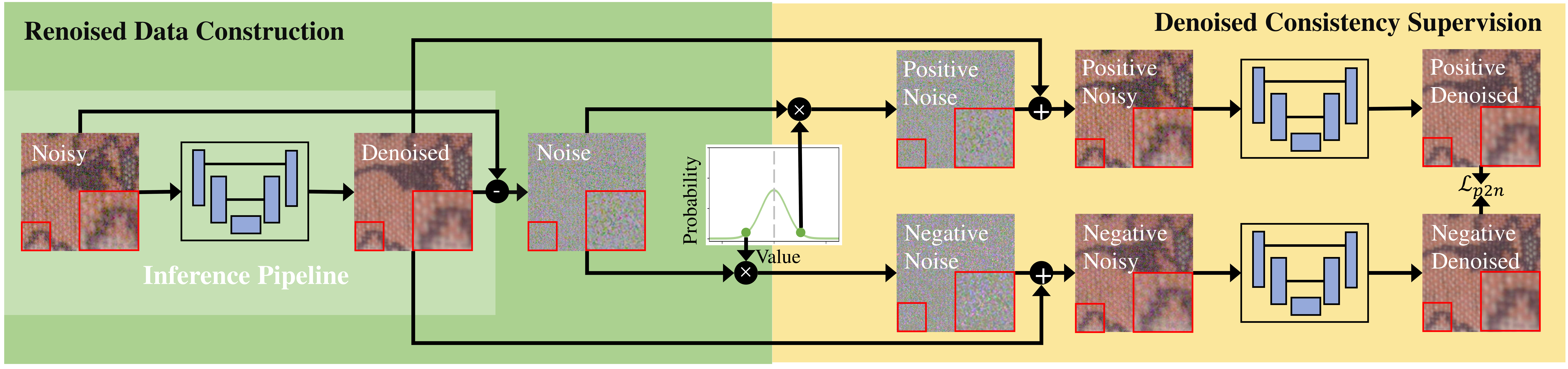} 
\caption{\textbf{Overview of the proposed self-supervised single image denoising paradigm.} 
	For training, the Positive2Negative paradigm consists of 2 steps. 
	The first step is Renoised Data Construction (RDC): multiply the predicted noise by positive and negative coefficients to construct multi-scale Positive noise and Negative noise, respectively. 
	The second step is Denoised Consistency Supervision (DCS): train the network through supervising the consistency of the denoised images.
	For inference, denoising can be achieved in just one pass through the network.
	(The networks shown in the figure are the same one. In addition, for better visualization, the images within the red boxes on the bottom left have been zoomed in and are displayed in the red boxes on the bottom right.)
}
	\label{fig:method}
\end{figure*}

\subsection{Motivation}
\label{sec:Motivation}

Our inspiration stems from the widely recognized assumptions that the noise distribution is zero-mean and approximately symmetrical~\cite{N2N, N2V}, which is consistent with our observation. 
In our observation experiment, we calculate the noise distribution of a fixed pixel across multiple observations of the same scene.
Specifically, following SIDD~\cite{SIDD}, ELD~\cite{ELD} and PMN~\cite{PMN}, we collect noisy image $y$ and clean image $c$. Then we calculate the noise value $n$ at each spatial location \((i, j)\). 
Additionally, we compute the frequency of $n_{i,j}$ cross noisy images of the same scene. 
The results are shown in Figure~\ref{fig:noise_distribution}, which demonstrates the noise distribution is zero-mean and approximately symmetrical.

Based on this observation, an intuitive insight is that the opposite noise $-n$ also follows the same distribution as that of the original noise $n$.
This intuitive idea inspires us to construct the opposite noise $n_{n}$ and correspoding opposite noisy image $y_{n}$:
\begin{align}
	n_{n} &= -{n}, \\
	y_{n} &= x + n_{n} = x - n.
\end{align}
The opposite noise $n_n$ and opposite noisy image $y_{n}$ are similar to the original noise $n$ and original noisy image $y$, respectively, as shown in Figure~\ref{fig:intuitive}. 

Based on the above observations and insights, we propose the Positive2Negative paradigm, as shown in Figure~\ref{fig:method}. 
Firstly, we extend our intuitive idea into a flexible and upgraded version: Renoised Data Construction (RDC), 
which constructs multiple positive and negative noises. 
Secondly, we propose the Denoised Consistency Supervision (DCS), which supervises the consistency of the predicted denoised images to learn a denoising neural network.

It is important to note that we do not require the noise distribution to be strictly zero-mean or symmetric. 
Although real noise generally has a zero-mean, certain cases where the the noise is not strictly  symmetric can be easily addressed by adjusting the norm in the loss function~\cite{N2N}.

\subsection{Renoised Data Construction (RDC)}
\label{sec:RDC}

The basic idea of RDC is to denoise the image to predict a denoised image and renoise the predicted denoised image by the predicted noise to construct multiple noisy images.

For a given noisy image $y=x+n$, where $x$ presents the clean image and $n$ presents the noise.
RDC firstly denoises the noisy image $y$ once through the neural  network $\mathcal{F}$ parametered by $\theta$:
\begin{equation}
	\hat{x} = \mathcal{F}_{\theta}(y).
\end{equation}
Here $\hat{x}$ is the predicted denoised image of the noisy observation $y$.
Then we can calculate the predicted noise $\hat{n}$ within the noisy observation $y$:
\begin{equation}
	\hat{n} = y-\hat{x}.
\end{equation}

Next, RDC constructs noisy images as training data based on the predicted noise $\hat{n}$ and the predicted denoised image $\hat{x}$.
As noise is zero-mean, we further design a zero-mean sampling strategy to construct new noises. Specifically, RDC multiplies the predicted noise $\hat{n}$ by both positive scale parameters $\sigma_p$ and negative scale parameters $-\sigma_n$ sampled from Gaussian distribution $\mathcal{N}$, constructing multi-scale Positive noise $n_{p}$ and Negative noise $n_{n}$, respectively:
\begin{align}
	\sigma_n &, \sigma_p \sim \mathcal{N}(1,\sigma), \\
	n_{p} &= \sigma_n \hat{n}, \\
	y_{p} &= \hat{x} + n_{p} = \hat{x} + \sigma_n\hat{n}, \\
	n_{n} &= -\sigma_p \hat{n}, \\
	y_{n} &= \hat{x} + n_{n} = \hat{x} - \sigma_p\hat{n}.
\end{align}

\begingroup
\renewcommand{\arraystretch}{1.2} 
\begin{table*}[h]
\caption{\textbf{Quantitative comparisons} on SIDD, CC, PolyU and FMDD datasets.
	 The best and second-best results (PSNR↑ / SSIM↑) are marked in {\color{red}red} and {\color{blue}blue} in self-supervised single image denoising methods. 
The results of noisy dataset based self-supervised methods and supervised methods are also provided, only as a reference comparison.
SIDD val and SIDD ben represent the SIDD validation dataset and the SIDD online benchmark, respectively.
}
	\label{tab:main}
	\small
	\setlength{\tabcolsep}{8.5pt}  
	\centering
	\begin{tabular}{c|lccccc}
		\toprule
		Category &Method &SIDD val~\cite{SIDD}&SIDD ben~\cite{SIDD} &CC~\cite{CC} &PolyU~\cite{PolyU} &FMDD~\cite{FMDD} \\ 
		\midrule

& \color{black}{Unprocessing~\cite{Unprocessing}}  
& \color{black}{28.12 / 0.529} & \color{black}{31.64 / 0.531}& \color{black}{35.18 / 0.908} &\color{black}{36.99 / 0.941} & \color{black}{28.45 / 0.574} \\	

		& \color{black}{MaskedDenoising~\cite{MaskedDenoising}}	
&  \color{black}{28.67 / 0.604} &  \color{black}{31.99 / 0.601}&  \color{black}{33.87 / 0.930} & \color{black}{34.56 / 0.936} &  \color{black}{29.86 / 0.643}  \\	
	
		\color{black}\multirow{-3}{*}{Supervised}
		&\color{black}{CLIPDenoising~\cite{CLIPDenoising}}    
		&\color{black} {34.79 / 0.866} & \color{black}{35.82 / 0.859}& \color{black}{36.30 / 0.941} &\color{black}{37.54 / 0.960} & \color{black}{30.56 / 0.698} \\		
		\midrule

		\color{black}{Self-Supervised} &\color{black}Noise2Void \cite{N2V} 
		& \color{black}29.35 / 0.651  & \color{black}{27.68 / 0.668} & \color{black}{32.27 / 0.862} & \color{black}{33.83 / 0.873} & {-}  \\

		\color{black}{(Noisy Dataset}   &  \color{black}CVF-SID~\cite{CVF-SID} 
		&\color{black}{34.81 / 0.872} & \color{black}{35.05 / 0.856} & \color{black}{29.10 /0.852}  & \color{black}{33.05 / 0.911} & \color{black}{31.64 / 0.837}  \\

		\color{black}{Based Methods)} & \color{black}LUD-VAE~\cite{LUD-VAE} 
		&\color{black}{34.91 / 0.892} & \color{black}{35.49 / 0.883} & \color{black}{31.25 / 0.920} & \color{black}{33.68 / 0.942} & {32.00 / 0.822}              \\
		\midrule
		
		&DIP~\cite{DIP} 
		&{32.11 / 0.740} & {-}             & {35.61 / 0.912} & {37.17 / 0.912} & {32.90 / 0.854}  \\ 
		
		&PD-denoising~\cite{PD} 
		&{33.97 / 0.820} & -               & {35.85 / 0.923} & {37.04 / 0.940} & {33.01 / 0.856}  \\
		
		&NN+denoiser~\cite{NN} 
		& {-}            & -               & {36.52 / 0.943} & {37.66 / 0.956} & {32.21 / 0.831}  \\
		
		&Self2Self~\cite{Self2self} 
		&{29.46 / 0.595} & {29.51 / 0.651} &{\color{blue}37.44 / 0.948} & {37.52 / 0.951} & {30.76 / 0.695}  \\
				
		{Self-Supervised}&R2R~\cite{R2R} 
		&{24.59 / 0.387} & {-}             & {33.43 / 0.865} & {36.23 / 0.931} & {27.17 / 0.526}  \\

		{(Single Image}&APBSN-single~\cite{APBSN} 
		&{30.90 / 0.818} & {-}             & {27.72 / 0.891} & {29.61 / 0.897} & {28.43 / 0.804}  \\
		
		{Based Methods)}&ScoreDVI~\cite{ScoreDVI} 
		&{34.75 /} {\color{blue}0.856} & {\color{blue}35.39 / 0.859} & {37.09 / 0.945} & {\color{blue}37.77 / 0.959} & {33.10 / 0.865}    \\  
				 
		&ZS-N2N~\cite{ZSN2N} 
		&{25.59 / 0.433} & {30.19 / 0.428} & {33.51 / 0.851} & {35.99 / 0.914} & {31.65 / 0.767} 
  		\\

		&MASH~\cite{MASH} 
	&{\color{blue}35.06} / 0.851 &34.80 / 0.814
&{31.17 / 0.889} & {37.62 / 0.932} &{\color{blue}33.71 / 0.882}   \\
		
		&ATBSN-single~\cite{ATBSN} 
		&32.39 / 0.796  &27.86 / 0.312               &31.87 / 0.897 &33.21 / 0.925 & 31.23 / 0.829 \\		
		\cline{2-7}

		&\textbf{Positive2Negative}

		&{\color{red}35.24 / 0.889} &{\color{red}35.80 / 0.875} 
		&{\color{red}37.92 / 0.954} &{\color{red}38.20 / 0.961} &{\color{red} 34.00 / 0.887} 

		\\

		\bottomrule
	\end{tabular}
\end{table*}
\endgroup

\subsection{Denoised Consistency Supervision (DCS)}
\label{sec:DCS}

The basic idea of DCS is that the noisy observations $y_{p}$ and $y_{n}$ should correspond to the same clean image. 
Therefore, the loss supervises the consistency of the predicted denoised images:
\begin{align}
	\mathcal{L}_{p2n} &=\| \mathcal{F}_{\theta}(y_p) - \mathcal{F}_{\theta}(y_n)   \| ,
\end{align}
where $\| \cdot \|$ represents the norm.
Considering the general nature of noise~\cite{N2N}, we employ a gradually varying norm function $\| \cdot \|_{2\rightarrow1.5}$ to cover a broader range of unknown distributions:
\begin{align}
	\| x \|_{2\rightarrow1.5} &= {( \left| x \right| + \epsilon )}^\gamma,
\end{align}
where $\epsilon=10^{-8}$ and $\gamma$ varies linearly from 2 to 1.5 during training.
The advantage of the varying 
$\| \cdot \|_{2\rightarrow1.5}$
norm is that $\| \cdot \|_{2\rightarrow1.5}$ does not require fine-tuning of parameters, yet $\| \cdot \|_{2\rightarrow1.5}$ consistently delivers robust performance.

The noisy observations $y_{p}$ and $y_{n}$ contain multi-scale noise. Through DCS, the neural network $\mathcal{F}_{\theta}$ gradually learns to become insensitive to input noise and consistently outputs the clean signal, achieving the desired denoising perfromance.

\subsection{Analysis}
\label{sec:Analysis}
In this section, we provide an analysis and proof demonstrating how the Positive2Negative paradigm achieves denoising.

Taking the 2-norm as an example, the loss function can be expressed as:
\begin{align}
	\mathcal{L}_{p2n} = \mathbb{E} \left[ \| \mathcal{F}_{\theta}(y_p) - \mathcal{F}_{\theta}(y_n) \|^2 \right].
\end{align}

Through Taylor expansion, we can approximate the predicted denoised images $\mathcal{F}_{\theta}(y_p)$ and $\mathcal{F}_{\theta}(y_n)$:
\begin{align}
	\mathcal{F}_{\theta}(y_p) &= \mathcal{F}_{\theta}\left(\mathcal{F}_{\theta}(y)\right) + \sigma_p \cdot \frac{\partial \mathcal{F}_{\theta}(y)}{\partial y} \cdot n + \mathcal{O}(n^2),\\
	\mathcal{F}_{\theta}(y_n) &= \mathcal{F}_{\theta}\left(\mathcal{F}_{\theta}(y)\right) - \sigma_n \cdot \frac{\partial \mathcal{F}_{\theta}(y)}{\partial y} \cdot n + \mathcal{O}(n^2).
\end{align}

Substituting the approximations, we get:
\begin{align}
	\mathcal{F}_{\theta}(y_p) - \mathcal{F}_{\theta}(y_n) \approx (\sigma_p+\sigma_n) \cdot \frac{\partial \mathcal{F}_{\theta}(y)}{\partial y} \cdot n.
\end{align}

With further Taylor expansion, the loss function can be simplified as:
\begin{align}
	\mathcal{L}_{p2n} 
	&\approx (4 + 2\sigma^2) \left\| \frac{\partial \mathcal{F}_{\theta}(x)}{\partial x} \cdot n \right\|^2.
	\label{eq:L_p2n_simplified}
\end{align}

Through consistency loss $L_{p2n}$, we optimize the network $\mathcal{F}_{\theta}$ such that $\frac{\partial \mathcal{F}_{\theta}(x)}{\partial x} \cdot n$ is optimized to drive towards zero, which means that the output $\mathcal{F}_{\theta}(y)$ tends to:
\begin{align}
	\mathcal{F}_{\theta}(y) &= \mathcal{F}_{\theta}(x + n),\\ 
	&= \mathcal{F}_{\theta}(x) + \frac{\partial \mathcal{F}_{\theta}(x)}{\partial x} \cdot n + \mathcal{O}(n^2),\\
	&\approx \mathcal{F}_{\theta}(x).
\end{align}

Ideally, the neural network $\mathcal{F}_{\theta}$ has learned image priors, the model output $\mathcal{F}_{\theta}(x)$ matches the clean signal $x$ and we have $\mathcal{F}_{\theta}(x) = x$.
Thus, the ultimate goal achieves:
\begin{align}
	\mathcal{F}_{\theta}(y) \approx x.
\end{align}

In summary, the neural network $\mathcal{F}_{\theta}$ gradually learns to denoise the noisy image $y$ effectively.

\begin{figure*}[h!]

\setkeys{Gin}{width=0.135\linewidth}
\captionsetup[subfigure]{skip=0.01ex,
belowskip=0.5ex,
labelformat=simple}
\renewcommand\thesubfigure{}
\setlength\tabcolsep{1.5pt}

\begin{tabular}{ccccccc}

\small{Noisy}  &\small {ScoreDVI~\cite{ScoreDVI} } & \small{ZS-N2N~\cite{ZSN2N}} & \small{MASH~\cite{MASH}} & \small {ATBSN-single~\cite{ATBSN}} & \small{\textbf{P2N (Ours)}} &\small {Reference}\\

{\includegraphics{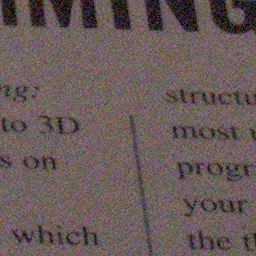}} &
{\includegraphics{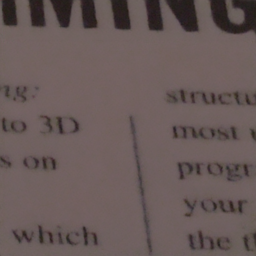}}& {\includegraphics{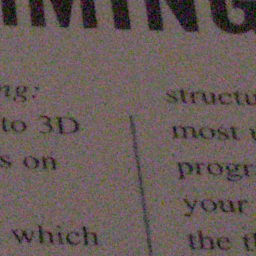}}& 
{\includegraphics{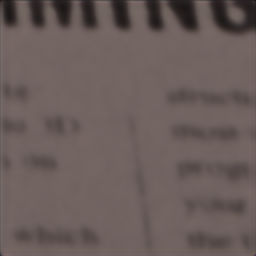}}& {\includegraphics{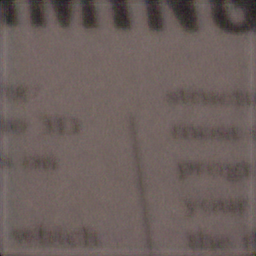}}& {\includegraphics{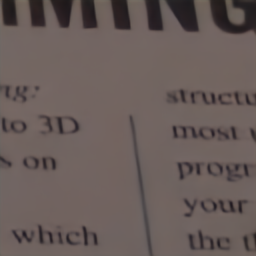}} &
{\includegraphics{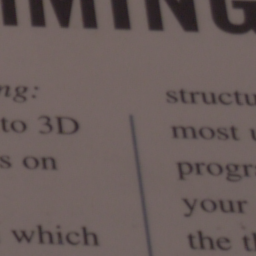}}  \\
\small25.54 / 0.432
& \small\color{blue}36.17 / 0.958
& \small27.10 / 0.529
& \small32.48 / 0.878
& \small30.23 / 0.795
& \small\color{red}37.67 / 0.970 
& \small PSNR / SSIM \\

{\includegraphics{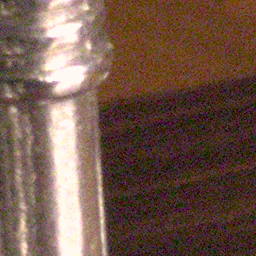}} &
{\includegraphics{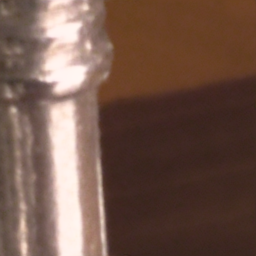}}& {\includegraphics{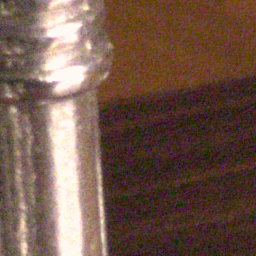}}& 
{\includegraphics{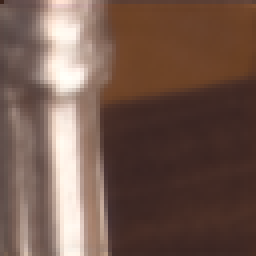}}& {\includegraphics{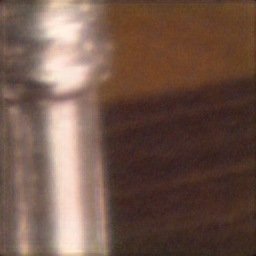}}& {\includegraphics{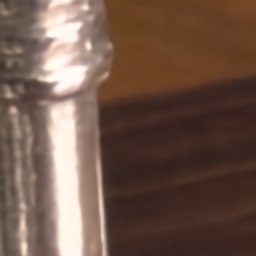}} &
{\includegraphics{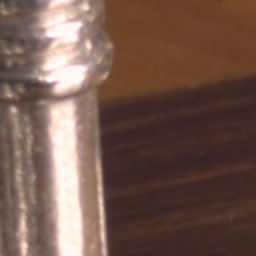}}  \\
\small 24.99 / 0.479
& \small \color{blue}34.32 / 0.930
& \small 26.63 / 0.556
& \small 30.64 / 0.881
& \small 27.12 / 0.799
& \small \color{red}35.14 / 0.950 
& \small PSNR / SSIM \\

{\includegraphics{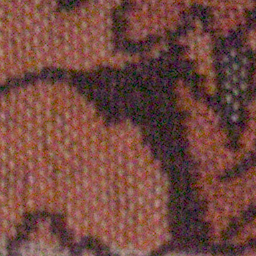}} &
{\includegraphics{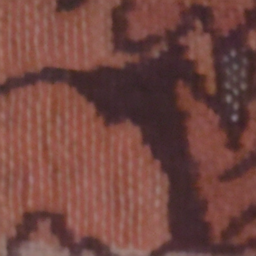}}& {\includegraphics{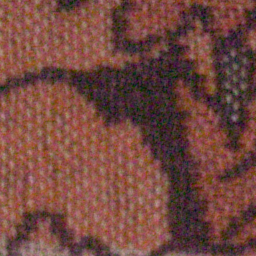}}& 
{\includegraphics{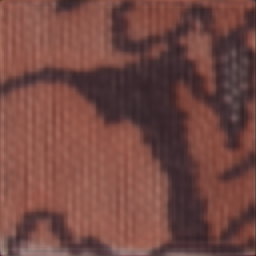}}& {\includegraphics{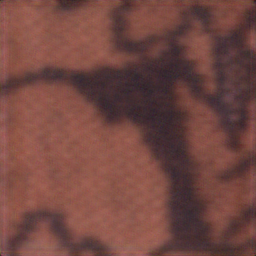}}& {\includegraphics{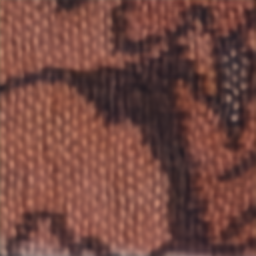}} &
{\includegraphics{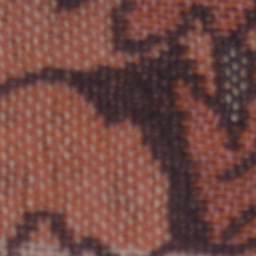}}  \\
\small 24.99 / 0.566
& \small \color{blue}32.12 / 0.848
& \small 26.52 / 0.638
& \small 31.23 / 0.791
& \small 28.22 / 0.643
& \small \color{red}32.95 / 0.887 
& \small PSNR / SSIM \\

\end{tabular}

\captionof{figure}
{\label{fig:main} 
\textbf{Qualitative comparisons} of P2N against other self-supervised single image denoising methods in SIDD validation dataset. The denoised results of ScoreDVI are always flaky, with discontinuous lines and loss of texture. ZS-N2N does not completely denoise the images. MASH and ATBSN-single tend to either meet aliasing effect or blur the details.
In summary, other methods meet aliasing effects and loss of texture, while Positive2Negative achieves both detailed texture preservation and comprehensive noise removal.
In the figure, P2N represents Positive2Negative.
}
\vspace{-1em}
\end{figure*}

\section{Experiments}

In this section, we first introduce the experimental settings in Section~\ref{sec:settings}. After that, we demonstrate the performance and comparisons with other methods in Section~\ref{sec:denoising_experiments}. Finally, we conduct a detailed ablation study in Section~\ref{sec:ablation}.
\subsection{Settings}
\label{sec:settings}
In this section, we briefly introduce the experimental settings and details. 

\noindent\textbf{Datasets details.} Following previous methods~\cite{MASH,ScoreDVI}, we evaluate the proposed paradigm on the widely-used datasets in self-supervised single image denoising: the SIDD validation dataset~\cite{SIDD}, the SIDD benchmark dataset~\cite{SIDD}, CC~\cite{CC}, PolyU~\cite{PolyU} and FMDD~\cite{FMDD}. The SIDD, PolyU and CC datasets consist of natural sRGB images, while the FMDD dataset consists of fluorescence microscopy images.

\noindent\textbf{Implementation details.} Following the default settings of self-supervised single image denoising methods~\cite{N2N, Blind2Unblind, MASH,ScoreDVI}, we employ the same neural network architecture as Noise2Noise~\cite{N2N} and MASH~\cite{MASH} and initialize the neural network parameters with those pre-trained on DIV2K dataset with Gaussian noise~\cite{ScoreDVI,NN}. For training details, we employ AdamW~\cite{AdamW} optimizer with a learning rate of $0.0001$ and set scale parameter $\sigma = 0.75$, for all the experiments. No image enhancement techniques are applied.

\noindent\textbf{Compared methods.} Following previous methods~\cite{ScoreDVI,MASH,DMID}, we mainly compare with self-supervised single image denoising methods.
In addition, we also present the results of noisy dataset based self-supervised and supervised image denoising methods only as a reference comparison. All the results of the comparison methods are sourced from previous work~\cite{ScoreDVI,MASH,DMID} or evaluated by ourselves. 

\begin{figure*}[t]
	\centering
	\small
	\renewcommand{\arraystretch}{1.5} 
	\setlength{\tabcolsep}{1.5pt}
	\scalebox{0.97}{
		\begin{tabular}[b]{c cc |c| c cc |c}

			\includegraphics[width=.12\textwidth]{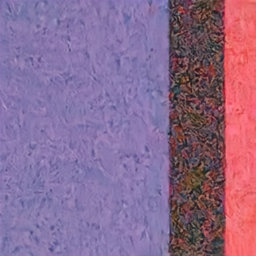} &   
			\includegraphics[width=.12\textwidth]{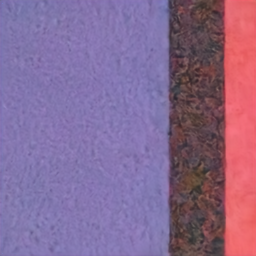} &
			\includegraphics[width=.12\textwidth]{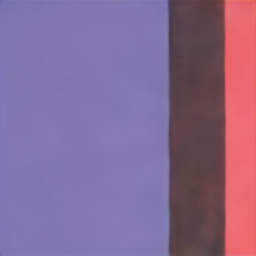} &
			\includegraphics[width=.12\textwidth]{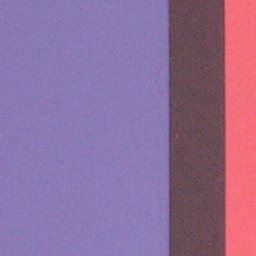} &
			
		\includegraphics[width=.12\textwidth]{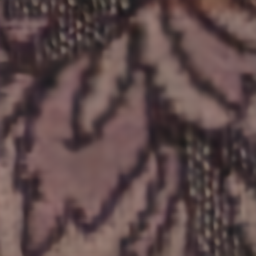} &   
		\includegraphics[width=.12\textwidth]{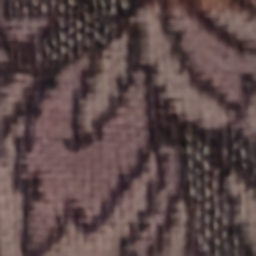} &
		\includegraphics[width=.12\textwidth]{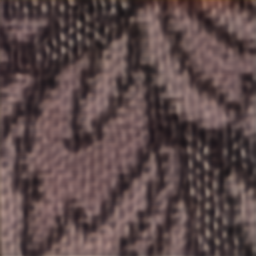} &
		\includegraphics[width=.12\textwidth]{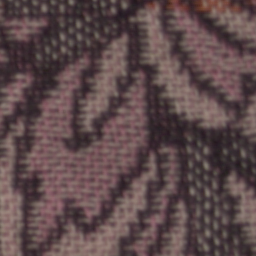} 
			\\
			\multicolumn{3}{c}{Denoised $\mathcal{F}_{\theta}(y)$} & Reference &\multicolumn{3}{c}{Denoised $\mathcal{F}_{\theta}(y)$} & Reference
			\\
		
			\includegraphics[width=.12\textwidth]{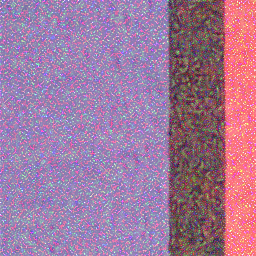} &   
			\includegraphics[width=.12\textwidth]{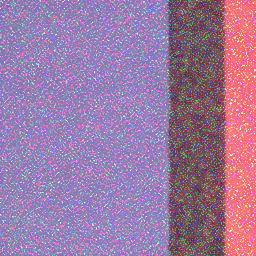} &
			\includegraphics[width=.12\textwidth]{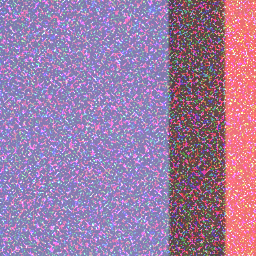} &
			\includegraphics[width=.12\textwidth]{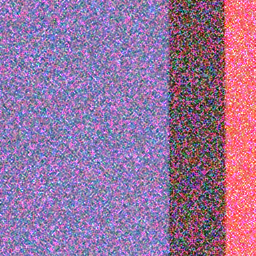} 
			
			&
			\includegraphics[width=.12\textwidth]{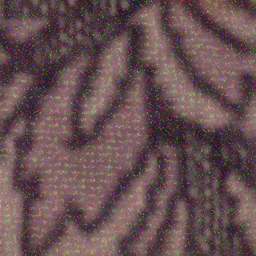} &   
			\includegraphics[width=.12\textwidth]{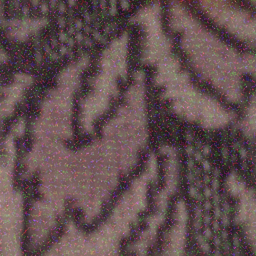} &
			\includegraphics[width=.12\textwidth]{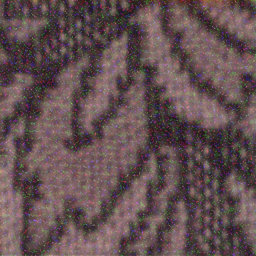} &
			\includegraphics[width=.12\textwidth]{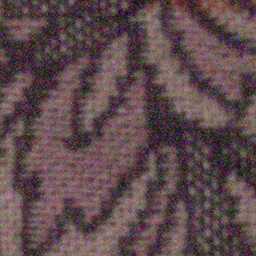} 
			\\
			\multicolumn{3}{c}{\makecell[c]{Reconstructed $y_n$}}& \multicolumn{1}{c}{\makecell[c]{Reference}}
			&
			\multicolumn{3}{c}{\makecell[c]{Reconstructed $y_n$}}& \multicolumn{1}{c}{\makecell[c]{Reference}}
			
			\\			
		\end{tabular}
	}
	\caption{\textbf{Visual results} of the training process. 
		The top row shows the denoised results, while the bottom row displays the reconstructed negative noisy images in the training process. On the far right of each set is the reference image.
		The denoised image and reconstructed noisy image gradually become more like the reference images.
	}
	\label{fig:ablation_training}
	\vspace{-1em}
\end{figure*}

\subsection{Comparisons}
\label{sec:denoising_experiments}

In this section, we demonstrate the performance of Positive2Negative and compare Positive2Negative with other methods both quantitatively and qualitatively.

\noindent\textbf{Quantitative comparisons.} The quantitative comparisons are described in Table \ref{tab:main}. 
Previous self-supervised single image denoising methods have shown varying performance across different datasets, but Positive2Negative consistently achieves better results. In fact, Positive2Negative even outperforms some dataset-based and supervised methods. Notably, in the online evaluation of the SIDD benchmark dataset, we achieve an improvement of over 0.4 dB.

\noindent\textbf{Qualitative comparisons.}
The qualitative comparisons are described in Figure~\ref{fig:main}. 
Clearly, the Positive2Negative paradigm achieves significant qualitative improvements by either enhancing details or effectively reducing noise. 
In contrast, other methods face limitations in both detail enhancement and noise removal due to the difficulty in learning denoising, resulting from the information-lossy barrier.


\subsection{Ablations}

\label{sec:ablation}
In this section, we first present the training process in Section~\ref{sec:training}. Then, we conduct ablation studies on the components and hyperparameters in Section \ref{sec:components} and Section \ref{sec:hyperparameters}, respectively.
\subsubsection{Ablation of training}
\label{sec:training}
We present visual results of the training process to illustrate the effectiveness of our Positive2Negative paradigm. 
As shown in Figure~\ref{fig:ablation_training}, throughout the training process, the denoised images and the reconstructed negative noisy images gradually converge towards their respective references.

\begin{table}[t]
	\centering
	\small
	\setlength{\tabcolsep}{10.5pt}  
	\caption{\textbf{Ablation of components.} Each step is crucial for Positive2Negative.
		N2N and N2V represent Noise2Noise and Noise2Void, respectively.
	}
	\label{tab:components}
	\begin{tabular}{l|c c | c c}
		\toprule
		{} & \multicolumn{2}{c}{CC~\cite{CC}} & \multicolumn{2}{c}{SIDD~\cite{SIDD}}\\
		\cline{2-5}
		
		\multirow{-2}{*}{Method} &  PSNR     & SSIM    & PSNR    & SSIM \\ \midrule
		Baseline                 &  35.40    & 0.927   & 33.50   & 0.849 \\
		\midrule
		RDC+N2N                  &  34.08    & 0.868  & 25.42  & 0.414\\
		RDC+N2V                  &  33.79    & 0.861  & 24.51  & 0.368\\
		N2N+DCS                  &  17.55    & 0.698  & 24.20  & 0.815\\
		N2V+DCS                  &  14.14    & 0.494  & 22.30  & 0.743\\
		\midrule
		RDC+DCS                  &  37.92    & 0.954   & 35.24   & 0.889\\
		\bottomrule
	\end{tabular}
\end{table}

\subsubsection{Ablation of components}
\label{sec:components}
We evaluate the effectiveness of each component of the Positive2Negative paradigm on the CC and SIDD datasets. RDC constructs precise training data and DCS trains robust denoising neural networks. 
Both RDC and DCS are crucial for Positive2Negative. Replacing either component would result in a decline in performance, as shown in Table~\ref{tab:components}.

\subsubsection{Ablation of hyperparameters}
\label{sec:hyperparameters}
Throughout the entire training process, the only adjustable hyperparameter is the scale parameter $\sigma$, which is set to $0.75$ for all experiments.
As shown in Table~\ref{tab:hyperparameters}, the performance remains similar for $\sigma$ values ranging from $0.25$ to $0.75$. The scale parameter \(\sigma\) introduces a dynamic and wide range of noise levels, while the “Fixed" setting corresponds to \(\sigma = 0\).

\begin{table}[t]
	\centering
	\setlength{\tabcolsep}{11pt}  
	\caption{\textbf{Ablation of hyperparameters.} The scale parameter is the only hyperparameter and is very robust.
		We fix the scale parameter $\sigma=0.75$ for all the experiments.
	}
	\small
	\label{tab:hyperparameters}
	\begin{tabular}{l|c c | c c}
		\toprule
		{} & \multicolumn{2}{c}{CC~\cite{CC}} & \multicolumn{2}{c}{SIDD~\cite{SIDD}}\\
		\cline{2-5}
		
		\multirow{-2}{*}{Method}   &  PSNR     & SSIM   & PSNR    & SSIM \\ \midrule
		Fixed                      &  37.33    & 0.952  & 35.24   & 0.892 \\
$\sigma = 0.25$            &  37.64    & 0.945  & 35.29   & 0.893 \\
$\sigma = 0.50$            &  37.84    & 0.952  & 35.30   & 0.891\\
$\sigma = 0.75$            &  37.92    & 0.954  & 35.24   & 0.889 \\
$\sigma = 1.00$            &  37.86    & 0.954  & 35.11   & 0.884\\
		\bottomrule
	\end{tabular}
\end{table}

\section{Discussions}
\label{sec:discussions}


\subsection{Distribution}

Different norms are suitable for different noise distributions~\cite{N2N}. For simplicity, we have uniformly adopted the \(\| \cdot \|_{2\rightarrow1.5}\) norm, as real-world noise is typically centered around a zero mean with slight fluctuations.  
For noise that remains nearly zero-mean, such as in PolyU~\cite{PolyU}, the \(\| \cdot \|_{2\rightarrow1.5}\) norm and the \(\| \cdot \|_{2}\) norm yield similar performance. 
For noise that exhibits slight deviations from zero mean, as in SIDD~\cite{SIDD}, the \(\| \cdot \|_{2\rightarrow1.5}\) norm can outperform the \(\| \cdot \|_{2}\) norm, as shown in Table~\ref{tab:Norm}. The slight increase in PSNR further suggests that the deviation remains minimal.

\subsection{Convergence}

\begin{figure}
	\centering
\includegraphics[width=0.99\linewidth]{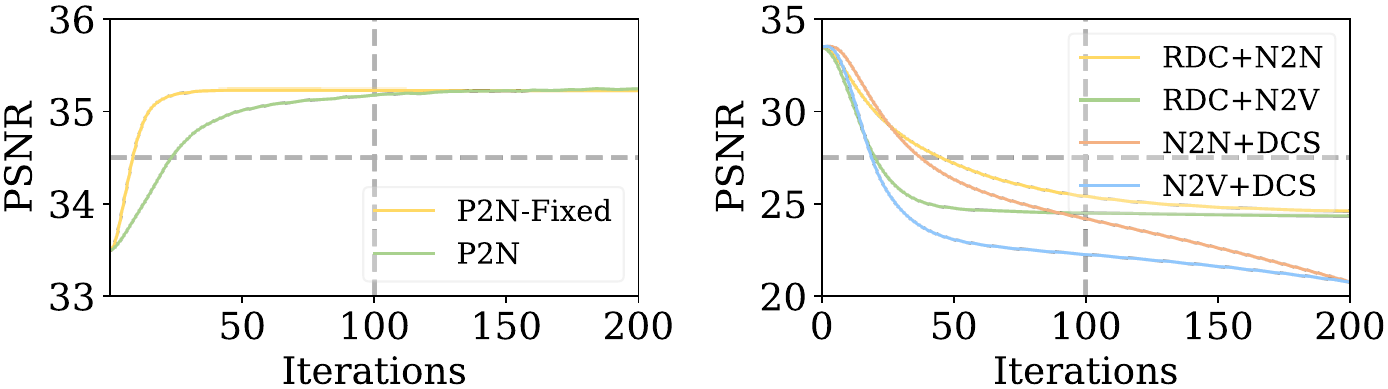} 
	\caption{\textbf{Convergence comparisons}.
		P2N converges quickly and remains stable after convergence on SIDD. The P2N-Fixed version (where $\sigma=0.00$), in particular, even achieves rapid convergence in approximately 10 iterations. P2N represents Positive2Negative.
	}
	\label{fig:Convergence}
\end{figure}

We present the PSNR variation over iterations on SIDD dataset~\cite{SIDD} in Figure~\ref{fig:Convergence}. 
Around the 100th iteration, the Positive2Negative paradigm converges and shows minimal fluctuations thereafter. 
Additionally, it is worth noting that the constructed data by RDC is unsuitable for Noise2Noise and Noise2Void. This is mainly because the noise in constructed images by RDC is correlated, which would quickly collapse the training processes of the other two methods.

\subsection{Efficiency}

For self-supervised single image denoising methods, computational efficiency is crucial, as each noisy image must be trained individually. Here, we demonstrate the number of training iterations, parameter, and computational cost, as detailed in Table~\ref{tab:Efficiency}.
It is obvious that our paradigm achieves significant efficiency improvements.

\begin{table}[t]
	\centering
	\small
	\setlength{\tabcolsep}{9.8pt}  
	\caption{\textbf{Discussion of distributions.} 
Different norms are suitable for different noise distributions~\cite{N2N}. For simplicity, we have uniformly adopted the \(\| \cdot \|_{2\rightarrow1.5}\) norm.
	}
	\label{tab:Norm}
	
	\begin{tabular}{l|c c c c}
		\toprule
		{Dataset} &$\| \cdot \|_{0}$&$\| \cdot \|_{1}$ &$\| \cdot \|_{2}$&$\| \cdot \|_{2\rightarrow1.5}$
		\\
		\midrule
		
PolyU~\cite{PolyU} &37.88 &38.02 &38.20 &38.20\\
SIDD~\cite{SIDD}    &35.12 &35.26 &35.20 &35.24\\
		\bottomrule
	\end{tabular}
\end{table}

\subsection{Comparison with Noise2Noise and Noise2Void}


Compared with Noise2Noise~\cite{N2N} and Noise2Void~\cite{N2V}, Positive2Negative also presents an elegant paradigm
The key differences are present in the following aspects.

Firstly, for the constructed data, the synthetic data constructed by Positive2Negative breaks the information-lossy barrier of Noise2Noise and Noise2Void. 
Secondly, for the supervision strategy, Positive2Negative enforces consistency between denoised results, whereas Noise2Noise and Noise2Void supervise the consistency between a denoised result and a noisy image.
Thirdly, from the assumption perspective, Noise2Noise, Noise2Void, and Positive2Negative all have theoretical guarantees, while Noise2Noise and Noise2Void operate under more restrictive assumptions. Noise2Noise requires two independent noisy images, while Noise2Void assumes that the noise is spatially correlated. 


\begin{table}[t]
	\centering
	\small
	\setlength{\tabcolsep}{8pt}  
	\caption{\textbf{Efficiency comparisons} 
		under input images with dimensions of 256$\times$256$\times$3.
	With similar parameters and FLOPs, P2N requires the fewest training iterations.
	P2N converges rapidly, and the P2N-Fixed version (where $\sigma = 0.00$) converges even just within 20 iterations. 
	P2N represents Positive2Negative.
	}
	\label{tab:Efficiency}
	
	\begin{tabular}{l | c c c}
		\toprule
		Method 
		& Iterations
		& Params (M)
		& FLOPs (G)
		
		\\ \midrule
		N2N~\cite{N2N} & - & 0.70 & 18.7\\
		B2U~\cite{Blind2Unblind} & - & 1.10 & 18.7\\
		\midrule
			
		Self2Self~\cite{Self2self}   & 450000     &1.00    &  9.6     \\
		R2R~\cite{R2R}                 & 8000        &0.56 & 36.7 \\
		ScoreDVI~\cite{ScoreDVI}        & 400       &13.5 & 37.9    \\
		MASH~\cite{MASH}                                &800 $\times$ 3   & 0.99       &  11.4     
		\\
		\midrule
		P2N-Fixed                                  &$<$20     & 0.99     & 11.4    
		\\
		P2N	                                       &$<$100     & 0.99     & 11.4    
		\\ \bottomrule
	\end{tabular}
\end{table}

\subsection{Differences with R2R and CVF-SID}


For the constructed data, R2R employs new Gaussian noise and applies reciprocal coefficients to the noisy image, while CVF-SID employs two networks to predict three components and applies fixed manual rules. In contrast, our method employs a single network to predict noise and applies random multiscale symmetrical parameters.

For the supervision strategy, R2R maps the denoised image to the noisy input, while CVF-SID primarily maps the first cyclic result to the noisy input and maps the second cyclic result to the first cyclic result. In contrast, our method maps the denoised image to another denoised image.

\subsection{Limitation and Future Work}


The primary limitation is that Positive2Negative has only been validated on the single image based self-supervised image denoising task. To prevent convergence to trivial solutions such as zero mapping or identity mapping, Positive2Negative requires training with a pre-trained Gaussian denoising model. While the single image based self-supervised denoising task allows training from a pre-trained model on Gaussian noise, the noisy dataset based self-supervised image denoising task trains from scratch. Therefore, extending the Positive2Negative paradigm to the noisy dataset based self-supervised image denoising task could be a promising direction for future work.

\section{Conclusion}

In this paper, we propose a self-supervised single image denoising paradigm, Positive2Negative, to break information-lossy barrier. 
Specifically, Positive2Negative includes two steps: RDC and DCS. 
Positive2Negative achieves SOTA performance compared to existing methods across various public benchmarks.

\section*{Acknowledgments}
This work is supported by the National Natural Science Foundation of China under Grants 62322204, 62131003, and the Key Laboratory of Target Cognition and Application Technology.

{
	\small
	\bibliographystyle{ieeenat_fullname}
	\bibliography{main}

\begin{thebibliography}{58}
\providecommand{\natexlab}[1]{#1}
\providecommand{\url}[1]{\texttt{#1}}
\expandafter\ifx\csname urlstyle\endcsname\relax
  \providecommand{\doi}[1]{doi: #1}\else
  \providecommand{\doi}{doi: \begingroup \urlstyle{rm}\Url}\fi

\bibitem[Abdelhamed et~al.(2018)Abdelhamed, Lin, and Brown]{SIDD}
Abdelrahman Abdelhamed, Stephen Lin, and Michael~S. Brown.
\newblock A high-quality denoising dataset for smartphone cameras.
\newblock In \emph{Proceedings of the IEEE/CVF Conference on Computer Vision
  and Pattern Recognition (CVPR)}, pages 1692--1700, 2018.

\bibitem[Batson and Royer(2019)]{Noise2self}
Joshua Batson and Loic Royer.
\newblock {Noise2Self}: Blind denoising by self-supervision.
\newblock In \emph{International Conference on Machine Learning (ICML)}, pages
  524--533, 2019.

\bibitem[Belthangady and Royer(2019)]{fluorescence_survey}
Chinmay Belthangady and Loic~A Royer.
\newblock Applications, promises, and pitfalls of deep learning for
  fluorescence image reconstruction.
\newblock In \emph{Nature Methods}, pages 1215--1225, 2019.

\bibitem[Brooks et~al.(2019)Brooks, Mildenhall, Xue, Chen, Sharlet, and
  Barron]{Unprocessing}
Tim Brooks, Ben Mildenhall, Tianfan Xue, Jiawen Chen, Dillon Sharlet, and
  Jonathan~T Barron.
\newblock Unprocessing images for learned raw denoising.
\newblock In \emph{Proceedings of the IEEE/CVF Conference on Computer Vision
  and Pattern Recognition (CVPR)}, pages 11036--11045, 2019.

\bibitem[Byun et~al.(2021)Byun, Cha, and Moon]{FBI}
Jaeseok Byun, Sungmin Cha, and Taesup Moon.
\newblock Fbi-denoiser: Fast blind image denoiser for poisson-gaussian noise.
\newblock In \emph{Proceedings of the IEEE/CVF Conference on Computer Vision
  and Pattern Recognition (CVPR)}, pages 5768--5777, 2021.

\bibitem[Chen et~al.(2023)Chen, Gu, Liu, Magid, Dong, Wang, Pfister, and
  Zhu]{MaskedDenoising}
Haoyu Chen, Jinjin Gu, Yihao Liu, Salma~Abdel Magid, Chao Dong, Qiong Wang,
  Hanspeter Pfister, and Lei Zhu.
\newblock Masked image training for generalizable deep image denoising.
\newblock In \emph{Proceedings of the IEEE/CVF Conference on Computer Vision
  and Pattern Recognition (CVPR)}, pages 1692--1703, 2023.

\bibitem[Chen et~al.(2022)Chen, Chu, Zhang, and Sun]{NAFNet}
Liangyu Chen, Xiaojie Chu, Xiangyu Zhang, and Jian Sun.
\newblock Simple baselines for image restoration.
\newblock In \emph{European Conference on Computer Vision (ECCV)}, pages
  17--33, 2022.

\bibitem[Chen et~al.(2024)Chen, Zhang, Yu, and Huang]{ATBSN}
Shiyan Chen, Jiyuan Zhang, Zhaofei Yu, and Tiejun Huang.
\newblock Exploring efficient asymmetric blind-spots for self-supervised
  denoising in real-world scenarios.
\newblock In \emph{Proceedings of the IEEE/CVF Conference on Computer Vision
  and Pattern Recognition (CVPR)}, pages 2814--2823, 2024.

\bibitem[Cheng et~al.(2023)Cheng, Liu, and Tan]{ScoreDVI}
Jun Cheng, Tao Liu, and Shan Tan.
\newblock Score priors guided deep variational inference for unsupervised
  real-world single image denoising.
\newblock In \emph{Proceedings of the IEEE/CVF International Conference on
  Computer Vision (ICCV)}, pages 12937--12948, 2023.

\bibitem[Cheng et~al.(2024)Cheng, Liang, and Tan]{CLIPDenoising}
Jun Cheng, Dong Liang, and Shan Tan.
\newblock Transfer clip for generalizable image denoising.
\newblock In \emph{Proceedings of the IEEE/CVF Conference on Computer Vision
  and Pattern Recognition (CVPR)}, pages 25974--25984, 2024.

\bibitem[Chihaoui and Favaro(2024)]{MASH}
Hamadi Chihaoui and Paolo Favaro.
\newblock Masked and shuffled blind spot denoising for real-world images.
\newblock In \emph{Proceedings of the IEEE/CVF Conference on Computer Vision
  and Pattern Recognition (CVPR)}, pages 3025--3034, 2024.

\bibitem[Feng et~al.(2024)Feng, Wang, Wang, Fan, and Huang]{PMN}
Hansen Feng, Lizhi Wang, Yuzhi Wang, Haoqiang Fan, and Hua Huang.
\newblock Learnability enhancement for low-light raw image denoising: A data
  perspective.
\newblock In \emph{IEEE Transactions on Pattern Analysis and Machine
  Intelligence (TPAMI)}, pages 370--387, 2024.

\bibitem[Hasinoff et~al.(2016)Hasinoff, Sharlet, Geiss, Adams, Barron, Kainz,
  Chen, and Levoy]{hasinoff2016photography}
Samuel~W Hasinoff, Dillon Sharlet, Ryan Geiss, Andrew Adams, Jonathan~T Barron,
  Florian Kainz, Jiawen Chen, and Marc Levoy.
\newblock Burst photography for high dynamic range and low-light imaging on
  mobile cameras.
\newblock In \emph{ACM Transactions on Graphics (ToG)}, pages 1--12, 2016.

\bibitem[Huang et~al.(2021)Huang, Li, Jia, Lu, and Liu]{Neighbor2neighbor}
Tao Huang, Songjiang Li, Xu Jia, Huchuan Lu, and Jianzhuang Liu.
\newblock {Neighbor2Neighbor}: Self-supervised denoising from single noisy
  images.
\newblock In \emph{Proceedings of the IEEE/CVF Conference on Computer Vision
  and Pattern Recognition (CVPR)}, pages 14781--14790, 2021.

\bibitem[Jang et~al.(2023)Jang, Lee, Park, Kim, and Cho]{CBSD}
Yeong~Il Jang, Keuntek Lee, Gu~Yong Park, Seyun Kim, and Nam~Ik Cho.
\newblock Self-supervised image denoising with downsampled invariance loss and
  conditional blind-spot network.
\newblock In \emph{Proceedings of the IEEE/CVF International Conference on
  Computer Vision (ICCV)}, pages 12196--12205, 2023.

\bibitem[Kim and Ye(2021)]{Noise2Score}
Kwanyoung Kim and Jong~Chul Ye.
\newblock Noise2score: tweedie’s approach to self-supervised image denoising
  without clean images.
\newblock In \emph{Advances in Neural Information Processing Systems}, pages
  864--874, 2021.

\bibitem[Kim et~al.(2022)Kim, Kwon, and Ye]{NDA}
Kwanyoung Kim, Taesung Kwon, and Jong~Chul Ye.
\newblock Noise distribution adaptive self-supervised image denoising using
  tweedie distribution and score matching.
\newblock In \emph{Proceedings of the IEEE/CVF Conference on Computer Vision
  and Pattern Recognition (CVPR)}, pages 1998--2006, 2022.

\bibitem[Krull et~al.(2019)Krull, Buchholz, and Jug]{N2V}
Alexander Krull, Tim-Oliver Buchholz, and Florian Jug.
\newblock Noise2void: Learning denoising from single noisy images.
\newblock In \emph{Proceedings of the IEEE/CVF Conference on Computer Vision
  and Pattern Recognition (CVPR)}, pages 2129--2137, 2019.

\bibitem[Krull et~al.(2020)Krull, Vi{\v{c}}ar, Prakash, Lalit, and Jug]{PN2V}
Alexander Krull, Tom{\'a}{\v{s}} Vi{\v{c}}ar, Mangal Prakash, Manan Lalit, and
  Florian Jug.
\newblock Probabilistic noise2void: Unsupervised content-aware denoising.
\newblock In \emph{Frontiers in Computer Science}, pages 2--5, 2020.

\bibitem[Laine et~al.(2019)Laine, Karras, Lehtinen, and Aila]{Laine}
Samuli Laine, Tero Karras, Jaakko Lehtinen, and Timo Aila.
\newblock High-quality self-supervised deep image denoising.
\newblock In \emph{Proceedings of International Conference on Neural
  Information Processing Systems (NeurIPS)}, pages 6970--6980, 2019.

\bibitem[Lee et~al.(2022)Lee, Son, and Lee]{APBSN}
Wooseok Lee, Sanghyun Son, and Kyoung~Mu Lee.
\newblock {AP-BSN}: Self-supervised denoising for real-world images via
  asymmetric pd and blind-spot network.
\newblock In \emph{Proceedings of the IEEE/CVF Conference on Computer Vision
  and Pattern Recognition (CVPR)}, pages 17725--17734, 2022.

\bibitem[Lehtinen et~al.(2018)Lehtinen, Munkberg, Hasselgren, Laine, Karras,
  Aittala, and Aila]{N2N}
Jaakko Lehtinen, Jacob Munkberg, Jon Hasselgren, Samuli Laine, Tero Karras,
  Miika Aittala, and Timo Aila.
\newblock {Noise2Noise}: Learning image restoration without clean data.
\newblock In \emph{International Conference on Machine Learning (ICML)}, pages
  4620--4631, 2018.

\bibitem[Li et~al.(2023{\natexlab{a}})Li, Zhang, Liu, Feng, Wang, Lei, and
  Zuo]{li2023spatially}
Junyi Li, Zhilu Zhang, Xiaoyu Liu, Chaoyu Feng, Xiaotao Wang, Lei Lei, and
  Wangmeng Zuo.
\newblock Spatially adaptive self-supervised learning for real-world image
  denoising.
\newblock In \emph{Proceedings of the IEEE/CVF Conference on Computer Vision
  and Pattern Recognition (CVPR)}, pages 9914--9924, 2023{\natexlab{a}}.

\bibitem[Li et~al.(2024)Li, Feng, Wang, Xiong, and Huang]{DMID}
Tong Li, Hansen Feng, Lizhi Wang, Zhiwei Xiong, and Hua Huang.
\newblock Stimulating the diffusion model for image denoising via adaptive
  embedding and ensembling.
\newblock In \emph{IEEE Transactions on Pattern Analysis and Machine
  Intelligence (TPAMI)}, pages 8240--8257, 2024.

\bibitem[Li et~al.(2023{\natexlab{b}})Li, Li, Zhou, Wu, Zhao, Fan, Deng, Wu,
  Xiao, He, Zhang, Zhang, Hu, Chen, Qiao, Xie, Li, Wang, Fang, and
  Dai]{DeepCADRT}
Xinyang Li, Yixin Li, Yiliang Zhou, Jiamin Wu, Zhifeng Zhao, Jiaqi Fan, Fei
  Deng, Zhaofa Wu, Guihua Xiao, Jing He, Yuanlong Zhang, Guoxun Zhang, Xiaowan
  Hu, Zhang~Yi Chen, Xingye~and, Hui Qiao, Hao Xie, Yulong Li, Haoqian Wang, Lu
  Fang, and Qionghai Dai.
\newblock Real-time denoising enables high-sensitivity fluorescence time-lapse
  imaging beyond the shot-noise limit.
\newblock In \emph{Nature Biotechnology}, page 282–292, 2023{\natexlab{b}}.

\bibitem[Loshchilov et~al.(2017)Loshchilov, Hutter, et~al.]{AdamW}
Ilya Loshchilov, Frank Hutter, et~al.
\newblock Fixing weight decay regularization in adam.
\newblock In \emph{arXiv preprint arXiv:1711.05101}, pages 1--10, 2017.

\bibitem[Mansour and Heckel(2023)]{ZSN2N}
Youssef Mansour and Reinhard Heckel.
\newblock {Zero-Shot Noise2Noise}: Efficient image denoising without any data.
\newblock In \emph{Proceedings of the IEEE/CVF Conference on Computer Vision
  and Pattern Recognition (CVPR)}, pages 14018--14027, 2023.

\bibitem[Moran et~al.(2020)Moran, Schmidt, Zhong, and Coady]{Noisier2noise}
Nick Moran, Dan Schmidt, Yu Zhong, and Patrick Coady.
\newblock {Noisier2Noise}: Learning to denoise from unpaired noisy data.
\newblock In \emph{Proceedings of the IEEE/CVF Conference on Computer Vision
  and Pattern Recognition (CVPR)}, pages 12064--12072, 2020.

\bibitem[Nam et~al.(2016)Nam, Hwang, Matsushita, and Kim]{CC}
Seonghyeon Nam, Youngbae Hwang, Yasuyuki Matsushita, and Seon~Joo Kim.
\newblock A holistic approach to cross-channel image noise modeling and its
  application to image denoising.
\newblock In \emph{Proceedings of the IEEE/CVF Conference on Computer Vision
  and Pattern Recognition (CVPR)}, pages 1683--1691, 2016.

\bibitem[Neshatavar et~al.(2022)Neshatavar, Yavartanoo, Son, and Lee]{CVF-SID}
Reyhaneh Neshatavar, Mohsen Yavartanoo, Sanghyun Son, and Kyoung~Mu Lee.
\newblock {CVF-SID}: Cyclic multi-variate function for self-supervised image
  denoising by disentangling noise from image.
\newblock In \emph{Proceedings of the IEEE/CVF Conference on Computer Vision
  and Pattern Recognition}, pages 17583--17591, 2022.

\bibitem[Nyquist(1928)]{Nyquist}
Harry Nyquist.
\newblock Certain topics in telegraph transmission theory.
\newblock In \emph{Transactions of the American Institute of Electrical
  Engineers}, pages 617--644, 1928.

\bibitem[Pan et~al.(2023)Pan, Liu, Liao, Cao, and Ren]{SDAP}
Yizhong Pan, Xiao Liu, Xiangyu Liao, Yuanzhouhan Cao, and Chao Ren.
\newblock Random sub-samples generation for self-supervised real image
  denoising.
\newblock In \emph{Proceedings of the IEEE/CVF International Conference on
  Computer Vision (ICCV)}, pages 12150--12159, 2023.

\bibitem[Pang et~al.(2021)Pang, Zheng, Quan, and Ji]{R2R}
Tongyao Pang, Huan Zheng, Yuhui Quan, and Hui Ji.
\newblock {Recorrupted-to-Recorrupted}: Unsupervised deep learning for image
  denoising.
\newblock In \emph{Proceedings of the IEEE/CVF Conference on Computer Vision
  and Pattern Recognition (CVPR)}, pages 2043--2052, 2021.

\bibitem[Papkov et~al.(2025)Papkov, Chizhov, and Parts]{SwinIA}
Mikhail Papkov, Pavel Chizhov, and Leopold Parts.
\newblock Swinia: Self-supervised blind-spot image denoising without
  convolutions.
\newblock In \emph{Proceedings of the Winter Conference on Applications of
  Computer Vision (WACV)}, pages 7071--7080, 2025.

\bibitem[Plotz and Roth(2017)]{DND}
Tobias Plotz and Stefan Roth.
\newblock Benchmarking denoising algorithms with real photographs.
\newblock In \emph{Proceedings of the IEEE/CVF Conference on Computer Vision
  and Pattern Recognition (CVPR)}, pages 1586--1595, 2017.

\bibitem[Quan et~al.(2020)Quan, Chen, Pang, and Ji]{Self2self}
Yuhui Quan, Mingqin Chen, Tongyao Pang, and Hui Ji.
\newblock {Self2Self} with dropout: Learning self-supervised denoising from
  single image.
\newblock In \emph{Proceedings of the IEEE/CVF Conference on Computer Vision
  and Pattern Recognition (CVPR)}, pages 1890--1898, 2020.

\bibitem[Shannon(1948)]{Shannon}
Claude~Elwood Shannon.
\newblock A mathematical theory of communication.
\newblock In \emph{The Bell System Technical Journal}, pages 379--423, 1948.

\bibitem[Ulyanov~Dmitry and Lempitsky(2018)]{DIP}
Andrea~Vedaldi Ulyanov~Dmitry and Victor Lempitsky.
\newblock Deep image prior.
\newblock In \emph{Proceedings of the IEEE/CVF Conference on Computer Vision
  and Pattern Recognition (CVPR)}, pages 9446--9454, 2018.

\bibitem[Wang et~al.(2020)Wang, Huang, Xu, Liu, Liu, and
  Wang]{wang2020photography}
Yuzhi Wang, Haibin Huang, Qin Xu, Jiaming Liu, Yiqun Liu, and Jue Wang.
\newblock Practical deep raw image denoising on mobile devices.
\newblock In \emph{European Conference on Computer Vision (ECCV)}, pages 1--16,
  2020.

\bibitem[Wang et~al.(2022)Wang, Liu, Li, and Han]{Blind2Unblind}
Zejin Wang, Jiazheng Liu, Guoqing Li, and Hua Han.
\newblock {Blind2Unblind}: Self-supervised image denoising with visible blind
  spots.
\newblock In \emph{Proceedings of the IEEE/CVF Conference on Computer Vision
  and Pattern Recognition (CVPR)}, pages 2027--2036, 2022.

\bibitem[Wang et~al.(2023)Wang, Fu, Liu, and Zhang]{LGBPN}
Zichun Wang, Ying Fu, Ji Liu, and Yulun Zhang.
\newblock {LG-BPN}: Local and global blind-patch network for self-supervised
  real-world denoising.
\newblock In \emph{Proceedings of the IEEE/CVF Conference on Computer Vision
  and Pattern Recognition (CVPR)}, pages 18156--18165, 2023.

\bibitem[Wei et~al.(2021)Wei, Fu, Zheng, and Yang]{ELD}
Kaixuan Wei, Ying Fu, Yinqiang Zheng, and Jiaolong Yang.
\newblock Physics-based noise modeling for extreme low-light photography.
\newblock In \emph{IEEE Transactions on Pattern Analysis and Machine
  Intelligence (TPAMI)}, pages 8520--8537, 2021.

\bibitem[Weigert et~al.(2018)Weigert, Schmidt, Boothe, M{\"u}ller, Dibrov,
  Jain, Wilhelm, Schmidt, Broaddus, Culley, et~al.]{CARE}
Martin Weigert, Uwe Schmidt, Tobias Boothe, Andreas M{\"u}ller, Alexandr
  Dibrov, Akanksha Jain, Benjamin Wilhelm, Deborah Schmidt, Coleman Broaddus,
  Si{\^a}n Culley, et~al.
\newblock Content-aware image restoration: pushing the limits of fluorescence
  microscopy.
\newblock In \emph{Nature Methods}, pages 1090--1097, 2018.

\bibitem[White(1990)]{learnability}
Halbert White.
\newblock Connectionist nonparametric regression: Multilayer feedforward
  networks can learn arbitrary mappings.
\newblock In \emph{Neural networks}, pages 535--549, 1990.

\bibitem[Wu et~al.(2006)Wu, Tan, Baird, DeCampo, White, and Wu]{PLMIC}
David Wu, Damian Tan, Marilyn Baird, John DeCampo, Chris White, and Hong~Ren
  Wu.
\newblock Perceptually lossless medical image coding.
\newblock In \emph{IEEE Transactions on Medical Imaging}, pages 335--344, 2006.

\bibitem[Wu et~al.(2012)Wu, Lin, and Karam]{PPDP}
Hong~Ren Wu, Weisi Lin, and Lina~J. Karam.
\newblock An overview of perceptual processing for digital pictures.
\newblock In \emph{IEEE International Conference on Multimedia and Expo
  Workshops}, pages 113--120, 2012.

\bibitem[Wu et~al.(2020)Wu, Liu, Cao, Ren, and Zuo]{D-BSN}
Xiaohe Wu, Ming Liu, Yue Cao, Dongwei Ren, and Wangmeng Zuo.
\newblock Unpaired learning of deep image denoising.
\newblock In \emph{European Conference on Computer Vision (ECCV)}, pages
  352--368, 2020.

\bibitem[Xie et~al.(2020)Xie, Wang, and Ji]{Noise2Same}
Yaochen Xie, Zhengyang Wang, and Shuiwang Ji.
\newblock Noise2same: Optimizing a self-supervised bound for image denoising.
\newblock In \emph{Advances in Neural Information Processing Systems}, pages
  20320--20330, 2020.

\bibitem[Xu et~al.(2018)Xu, Li, Liang, Zhang, and Zhang]{PolyU}
Jun Xu, Hui Li, Zhetong Liang, David Zhang, and Lei Zhang.
\newblock Real-world noisy image denoising: A new benchmark.
\newblock In \emph{arXiv preprint arXiv:1804.02603}, pages 1--13, 2018.

\bibitem[Xu et~al.(2020)Xu, Huang, Cheng, Liu, Zhu, Xu, and Shao]{NAC}
Jun Xu, Yuan Huang, Ming-Ming Cheng, Li Liu, Fan Zhu, Zhou Xu, and Ling Shao.
\newblock {Noisy-as-Clean}: Learning self-supervised denoising from corrupted
  image.
\newblock In \emph{IEEE Transactions on Image Processing (TIP)}, pages
  9316--9329, 2020.

\bibitem[Zamir et~al.(2022)Zamir, Arora, Khan, Hayat, Khan, and
  Yang]{Restormer}
Syed~Waqas Zamir, Aditya Arora, Salman Khan, Munawar Hayat, Fahad~Shahbaz Khan,
  and Ming-Hsuan Yang.
\newblock Restormer: Efficient transformer for high-resolution image
  restoration.
\newblock In \emph{Proceedings of the IEEE/CVF Conference on Computer Vision
  and Pattern Recognition (CVPR)}, pages 5728--5739, 2022.

\bibitem[Zhang et~al.(2023)Zhang, Zhou, Jiang, and Fu]{MMBSN}
Dan Zhang, Fangfang Zhou, Yuwen Jiang, and Zhengming Fu.
\newblock Mm-bsn: Self-supervised image denoising for real-world with
  multi-mask based on blind-spot network.
\newblock In \emph{Proceedings of the IEEE/CVF Conference on Computer Vision
  and Pattern Recognition Workshops}, pages 4189--4198, 2023.

\bibitem[Zhang et~al.(2019)Zhang, Zhu, Nichols, Wang, Zhang, Smith, and
  Howard]{FMDD}
Yide Zhang, Yinhao Zhu, Evan Nichols, Qingfei Wang, Siyuan Zhang, Cody Smith,
  and Scott Howard.
\newblock A poisson-gaussian denoising dataset with real fluorescence
  microscopy images.
\newblock In \emph{Proceedings of the IEEE/CVF Conference on Computer Vision
  and Pattern Recognition (CVPR)}, pages 11710--11718, 2019.

\bibitem[Zhang et~al.(2022)Zhang, Li, Law, Wang, Qin, and Li]{IDR}
Yi Zhang, Dasong Li, Ka~Lung Law, Xiaogang Wang, Hongwei Qin, and Hongsheng Li.
\newblock {IDR}: Self-supervised image denoising via iterative data refinement.
\newblock In \emph{Proceedings of the IEEE/CVF Conference on Computer Vision
  and Pattern Recognition (CVPR)}, pages 2098--2107, 2022.

\bibitem[Zheng et~al.(2020)Zheng, Tan, Zhang, Shi, Ma, and Bao]{NN}
Dihan Zheng, Sia~Huat Tan, Xiaowen Zhang, Zuoqiang Shi, Kaisheng Ma, and
  Chenglong Bao.
\newblock An unsupervised deep learning approach for real-world image
  denoising.
\newblock In \emph{International Conference on Learning Representations
  (ICLR)}, pages 1--9, 2020.

\bibitem[Zheng et~al.(2022)Zheng, Zhang, Ma, and Bao]{LUD-VAE}
Dihan Zheng, Xiaowen Zhang, Kaisheng Ma, and Chenglong Bao.
\newblock Learn from unpaired data for image restoration: A variational bayes
  approach.
\newblock In \emph{IEEE Transactions on Pattern Analysis and Machine
  Intelligence (TPAMI)}, pages 5889--5903, 2022.

\bibitem[Zhou et~al.(2020)Zhou, Jiao, Huang, Wang, Wang, Shi, and Huang]{PD}
Yuqian Zhou, Jianbo Jiao, Haibin Huang, Yang Wang, Jue Wang, Honghui Shi, and
  Thomas Huang.
\newblock When awgn-based denoiser meets real noises.
\newblock In \emph{Proceedings of the AAAI Conference on Artificial
  Intelligence (AAAI)}, pages 13074--13081, 2020.

\bibitem[Zou et~al.(2023)Zou, Yan, and Fu]{DCD}
Yunhao Zou, Chenggang Yan, and Ying Fu.
\newblock Iterative denoiser and noise estimator for self-supervised image
  denoising.
\newblock In \emph{Proceedings of the IEEE/CVF Conference on Computer Vision
  and Pattern Recognition (CVPR)}, pages 13265--13274, 2023.

\end{thebibliography}
}
\end{document}